\documentclass[10pt,twocolumn,letterpaper]{article}

\usepackage{cvpr}
\usepackage{times}
\usepackage{epsfig}
\usepackage{graphicx}
\usepackage{amsmath}
\usepackage{amssymb}
\usepackage{subcaption}
\usepackage{fixltx2e}
\usepackage{multirow}
\usepackage{gensymb}

\usepackage[pagebackref=true,breaklinks=true,letterpaper=true,colorlinks,bookmarks=false]{hyperref}

\cvprfinalcopy 

\def\cvprPaperID{156} 

\ifcvprfinal\fi
\begin{document}

\title{Deep Neural Networks predict Hierarchical Spatio-temporal Cortical Dynamics of Human Visual Object Recognition}

\author{Radoslaw M. Cichy$^{\dagger *}$\hspace*{5mm}Aditya Khosla$^\dagger$\hspace*{5mm}Dimitrios Pantazis$^\dagger$\hspace*{5mm}Antonio Torralba$^\dagger$\hspace*{5mm}Aude Oliva$^\dagger$\\
$^\dagger$Massachusetts Institute of Technology\hspace*{8mm}$^*$Free University Berlin\\
{\tt\small \{rmcichy, khosla, pantazis, torralba, oliva\}@mit.edu}
}

\maketitle

\begin{abstract}
The complex multi-stage architecture of cortical visual pathways provides the neural basis for efficient visual object recognition in humans. However, the stage-wise computations therein remain poorly understood. Here, we compared temporal (magnetoencephalography) and spatial (functional MRI) visual brain representations with representations in an artificial deep neural network (DNN) tuned to the statistics of real-world visual recognition. We showed that the DNN captured the stages of human visual processing in both time and space from early visual areas towards the dorsal and ventral streams. Further investigation of crucial DNN parameters revealed that while model architecture was important, training on real-world categorization was necessary to enforce spatio-temporal hierarchical relationships with the brain. Together our results provide an algorithmically informed view on the spatio-temporal dynamics of visual object recognition in the human visual brain.
\end{abstract}

\section{Introduction}
Visual object recognition in humans is mediated by complex multi-stage processing of visual information emerging rapidly in a distributed network of cortical regions~\cite{goodale1982analysis,felleman1991distributed,bullier2001integrated,milner1995visual,kourtzi2011neural,kravitz2011new,dicarlo2012does}. Understanding visual object recognition in cortex thus requires a predictive and quantitative model that captures the complexity of the underlying spatio-temporal dynamics~\cite{riesenhuber1999hierarchical,riesenhuber2002neural,naselaris2009bayesian}.

A major impediment in creating such a model is the highly nonlinear and sparse nature of neural tuning properties in mid- and high-level visual areas~\cite{david2006spectral,wang1996optical,yamane2008neural} that is difficult to capture experimentally, and thus unknown. Previous approaches to modeling object recognition in cortex relied on extrapolation of principles from well understood lower visual areas such as V1~\cite{riesenhuber1999hierarchical,riesenhuber2002neural} and strong manual intervention, achieving only modest task performance compared to humans.

Here we take an alternative route, constructing and comparing against brain signals a visual computational model based on deep neural networks (DNNs)~\cite{lecun2015deep,mnih2015human} , i.e., computer vision models in which model neuron tuning properties are set by supervised learning without manual intervention~\cite{lecun2015deep,rumelhart1988learning}. DNNs are the best performing models on computer vision object recognition benchmarks and yield human performance levels on object categorization~\cite{russakovsky2014imagenet,he2015delving}. We used a tripartite strategy to reveal the spatio-temporal processing cascade underlying human visual object recognition by DNN model comparisons.

First, as object recognition is a process rapidly unfolding over time~\cite{bullier2001integrated,cichy2014resolving,schmolesky1998signal}, we compared DNN visual representations to millisecond resolved magnetoencephalography (MEG) brain data. Our results delineate, to our knowledge for the first time, an ordered relationship between the stages of processing in computer vision model and the time course with which object representations emerge in the human brain.

Second, as object recognition recruits a multitude of distributed brain regions, a full account of object recognition needs to go beyond the analysis of a few pre-defined brain regions~\cite{agrawal2014pixels,cadieu2014deep,gucclu2014deep,khaligh2014deep,yamins2014performance}, determining the relationship between DNNs and the whole brain. Using a spatially unbiased approach, we revealed a hierarchical relationship between DNNs and the processing cascade of both the ventral and dorsal visual pathway.

Third, interpretation of a DNN-brain comparison depends on the factors shaping the DNN fundamentally: the pre-specified model architecture, the training procedure, and the learned task (e.g. object categorization). By comparing different DNN models to brain data, we demonstrated the influence of each of these factors on the emergence of similarity relations between DNNs and brains in both space and time.

Together, our results provide an algorithmically informed perspective of the spatio-temporal dynamics underlying visual object recognition in the human brain.

\section{Results}
\subsection{Construction of a DNN performing at human level in object categorization}
\begin{figure*}[t]
\begin{center}
   \includegraphics[width=0.9\linewidth]{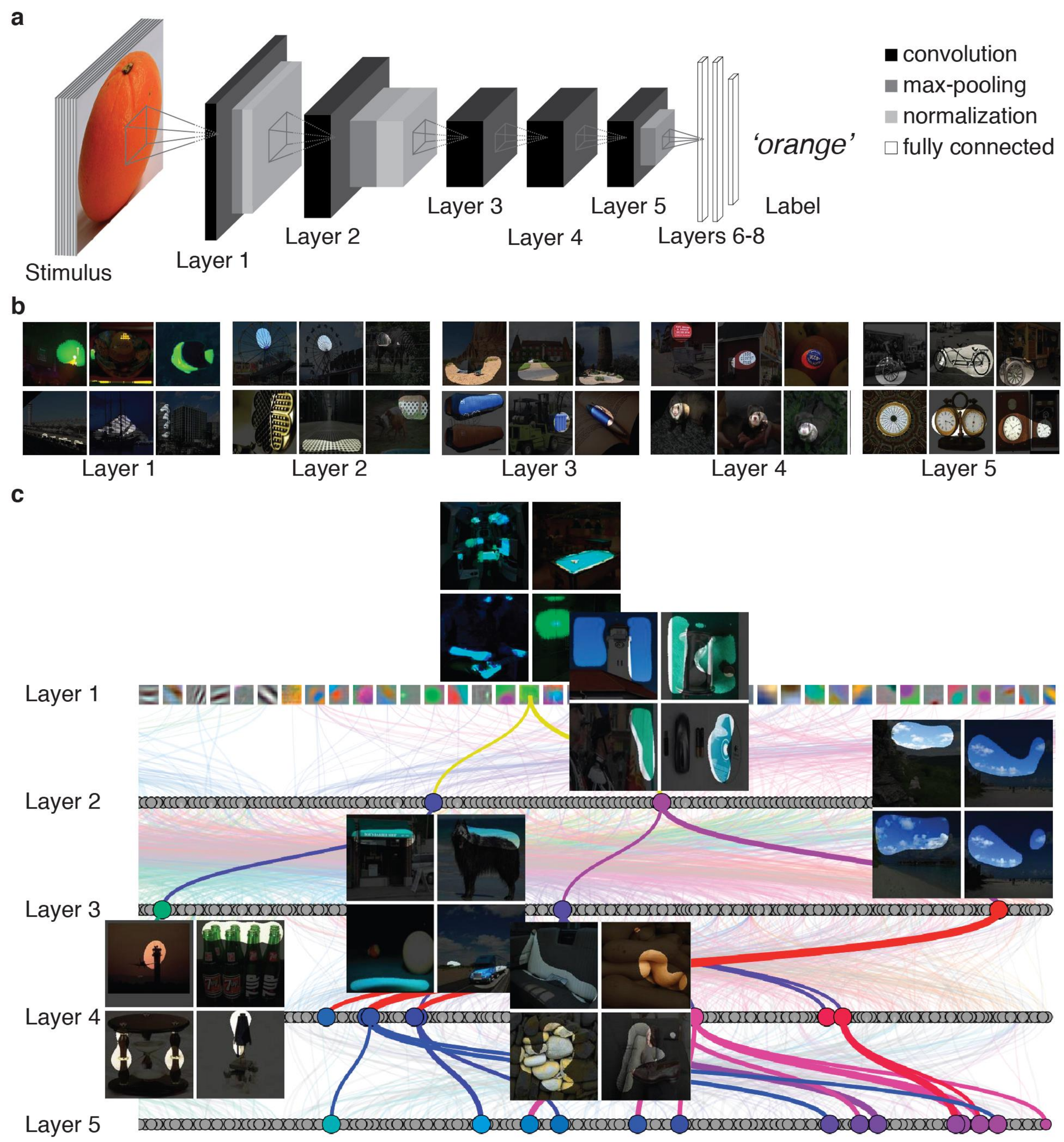}
\end{center}
\vspace*{-4mm}
\caption{\textbf{Deep neural network architecture and properties. (a)} The DNN architecture comprised 8 layers. Each of layers $1-5$ contained a combination of convolution, max-pooling and normalization stages, whereas the last three layers were fully connected. The DNN takes pixel values as inputs and propagates information feed-forward through the layers, activating model neurons with particular activation values successively at each layer. \textbf{(b)} Visualization of model receptive fields (RFs) selectivity. Each row shows the 4 images most strongly activating two exemplary model neurons for layers 1 through 5, with shaded regions highlighting the image area primarily driving the neuron response. \textbf{(c)} Visualization of example DNN connections and neuron RF selectivity. The thickness of highlighted lines (colored to ease visualization) indicates the weight of the strongest connections going in and out of neurons, starting from a sample neuron in layer 1. Combined visualization of neuron RF selectivity and connections between neurons, here starting from a sample neuron in layer 1 (only parts of the network for visualization). Neurons in layer 1 are represented by their filters, and in layers $2-5$ by gray dots. Inlays show the 4 images that most strongly activate each neuron. A complete visualization of all neurons in layers 1 through 5 is available at \href{http://brainmodels.csail.mit.edu}{http://brainmodels.csail.mit.edu}.}
\label{fig:fig1}
\end{figure*}

To be a plausible model of object recognition in cortex, a computational model must provide high performance on visual object categorization. Latest generations of computer vision models, termed deep neural networks (DNNs), have achieved extraordinary performance, thus raising the question whether their algorithmic representations bear resemblance of the neural computations underlying human vision. To investigate we created an 8-layer DNN architecture (Fig.~\ref{fig:fig1}(a)) that corresponds to the best-performing model in object classification in the ImageNet Large Scale Visual Recognition Challenge 2012~\cite{krizhevsky2012imagenet}. Each DNN layer performs simple operations that are implementable in biological circuits, such as convolution, pooling and normalization. We trained the DNN to perform object categorization on everyday object categories (683 categories, with ~1300 images in each category) using back propagation, i.e., the network learned neuronal tuning functions by itself. We termed this neural network object deep neural network (object DNN). The object DNN performed equally well on object categorization as previous implementations (Suppl.~Table~1). We investigated the coding of visual information in the object DNN by determining the receptive field (RF) selectivity of the model neurons using a neuroscience-inspired reduction method~\cite{scenecnn_iclr15}.

We found that neurons in early layers had Gabor filter or color patch-like sensitivity, while those of deeper layers had larger RFs and sensitivity to complex forms (Fig.~\ref{fig:fig1}(b)). Thus the object DNN learned representations in a hierarchy of increasing complexity, akin to representations in the primate visual brain hierarchy~\cite{kourtzi2011neural,dicarlo2012does}. Figure~\ref{fig:fig1}(c) exemplifies the connectivity and receptive field selectivity of the most strongly connected neurons starting from a sample neuron in layer 1. An online tool offering visualization of RF selectivity of all neurons in layers 1 through 5 is available at \href{http://brainmodels.csail.mit.edu}{http://brainmodels.csail.mit.edu}.

\subsection{Representational similarity analysis was used as the integrative framework for DNN-brain comparison}
To compare representations in the object DNN and human brains, we used a 118-image set of natural objects on real-world backgrounds (Fig.~\ref{fig:fig2}(a)). Note that these 118 images were not used for training the object DNN to avoid circular inference. With $94\%$ correct performance in a top-five categorization task on this 118 image set, the network performed at a level comparable to humans~\cite{russakovsky2014imagenet} (voting on each of the 118 images is available at \href{http://brainmodels.csail.mit.edu}{http://brainmodels.csail.mit.edu}).

We also recorded fMRI and MEG in 15 participants viewing random sequences of the same 118 real-world object image set while conducting an orthogonal task. The experimental design was adapted to the specifics of the measurement technique (Suppl.~Fig.~1).

We compared fMRI and MEG brain measurements with the DNN in a common analysis framework with representational similarity analysis~\cite{kriegeskorte2008representational} (Fig.~\ref{fig:fig2}(b)). The basic idea is that if two images are similarly represented in the brain, they should also be similarly represented in the DNN. To quantify, we first obtained signal measurements in temporally specific MEG sensor activation patterns (1ms steps from $-100$ to $+1000ms$), in spatially specific fMRI voxel patterns, and in layer-specific model neuron activations of the DNN. To make the different signal spaces (fMRI, MEG, DNN) comparable, we abstracted signals to a similarity space. In detail, for each signal space we computed dissimilarities ($1-$Spearman's $\rho$ for DNN and fMRI, percent decoding accuracy in pair-wise classification for MEG) between every pair of conditions (images), as exemplified by images 1 and 2 in Fig.~\ref{fig:fig2}(b).  This yielded $118 \times 118$ representational dissimilarity matrices (RDMs) indexed in rows and columns by the compared conditions. These RDMs were time-resolved for MEG, space-resolved for fMRI, and layer-resolved in DNN. Comparing DNN RDMs with MEG RDMs resulted in time courses highlighting how DNN predicted emerging visual representations. Comparing DNN RDMs with fMRI RDMs resulted in spatial maps indicative of how the object DNN predicted brain activity.

\subsection{The object DNN predicted temporal dynamics of emerging visual representations in the human brain}
Visual information processing in the brain is a process that rapidly evolves over time~\cite{bullier2001integrated,cichy2014resolving,schmolesky1998signal}, and a model of object recognition in cortex should mirror this temporal evolution. While the DNN used here does not model time, it has a clear sequential structure: information flows from one layer to the next in strict order. We thus investigated whether the object DNN predicted emerging visual representations in the first few hundred milliseconds of vision in sequential order. For this we determined representational similarity between layer-specific DNN representations and MEG data in millisecond steps from $-100$ to $+1000ms$ with respect to image onset and layer-specific DNN representations. We found that all layers of the object DNN were representationally similar to human brain activity, indicating that the model captures emerging brain visual representations (Fig.~\ref{fig:fig3}(a), $P < 0.05$ cluster definition threshold, $P < 0.05$ cluster threshold, lines above data curves color-coded same as those indicate significant time points, for details see Suppl.~Table~2). We next investigated whether the hierarchy of the layered architecture of the object DNN, as characterized by an increasing size and complexity of model RFs feature selectivity, corresponded to the hierarchy of temporal processing in the brain. That is, we examined whether low and high layers of the object DNN predicted early and late brain representations, respectively. We found this to be the case: There was a positive hierarchical relationship ($n = 15$, Spearman's $\rho = 0.35$, $P = 0.0007$) between the layer number of the object DNN and position in the hierarchy of the deep object network and the peak latency of the correlation time courses between object DNN and MEG RDMs and deep object network layer RDMs (Fig.~\ref{fig:fig3}(b)).

Together these analyses established, to our knowledge for the first time, a correspondence in the sequence of processing steps of a computational model of vision and the time course with which visual representations emerge in the human brain.

\subsection{The object DNN predicted the hierarchical topography of visual representations in the human ventral and dorsal visual streams}

To localize visual representations common to brain and the object DNN, we used a spatially unbiased surface-based searchlight approach. Comparison of representational similarities between fMRI data and object DNN RDMs yielded 8 layer-specific spatial maps identifying the cortical regions where the object DNN predicted brain activity (Fig.~\ref{fig:fig4}, cluster definition threshold $P < 0.05$, cluster-threshold $P < 0.05$; different viewing angles available in Suppl.~Movie~1).

The results indicate a hierarchical correspondence between model network layers and the human visual system. For low DNN layers, similarities of visual representations were confined to the occipital lobe, i.e., low- and mid-level visual regions, and for high DNN layers in more anterior regions in both the ventral and dorsal visual stream. A supplementary volumetric searchlight analysis (Suppl.~Text 1, Suppl.~Fig.~2; using a false discovery rate correction allowing voxel-wise inference reproduced these findings, yielding corroborative evidence across analysis methodologies.

These results suggest that hierarchical systems of visual representations emerge in both the human ventral and dorsal visual stream as the result of task constraints of object categorization posed in everyday life, and provide strong evidence for object representations in the dorsal stream independent of attention or motor intention.

\begin{figure*}[t]
\begin{center}
   \includegraphics[width=0.83\linewidth]{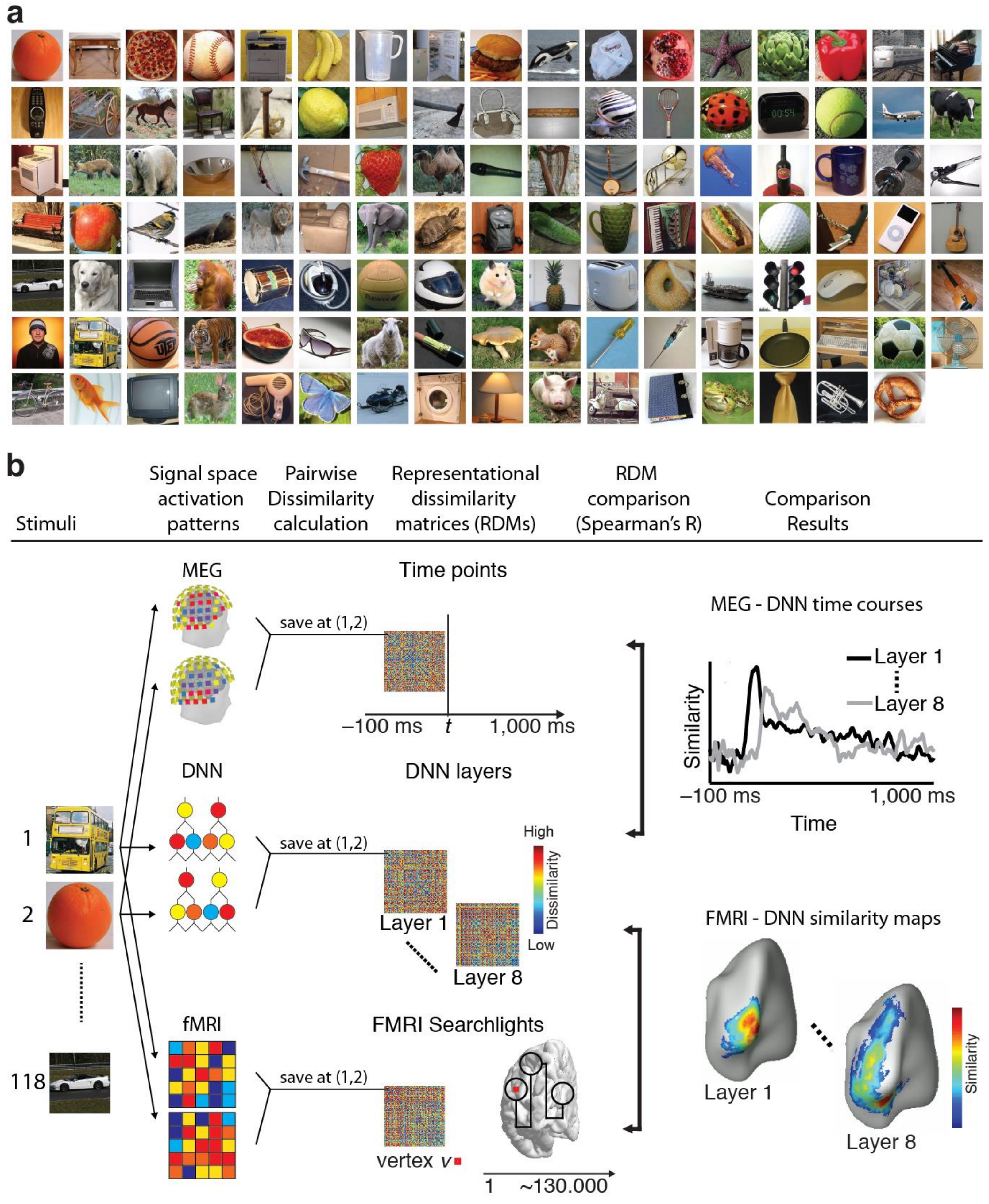}
\end{center}
\vspace*{-4mm}
\caption{\textbf{Stimulus set and comparison of brain and DNN representations. (a)} The stimulus set consisted of 118 images of distinct object categories. \textbf{(b)} Representational similarity analysis between MEG, fMRI and DNN data. In each signal space (fMRI, MEG, DNN) we summarized representational structure by calculating the dissimilarity between activation patterns of different pairs of conditions (here exemplified for two objects: bus and orange). This yielded representational dissimilarity matrices (RDMs) indexed in rows and columns by the compared conditions. We calculated millisecond resolved MEG RDMs from $–100ms$ to $+1000ms$ with respect to image onset, layer-specific DNN RDMs (layers 1 through 8) and voxel-specific fMRI RDMs in a spatially unbiased cortical surface-based searchlight procedure. RDMs were directly comparable (Spearman's $\rho$), facilitating integration across signal spaces. Comparison of DNN with MEG RDMs yielded time courses of similarity between emerging visual representations in the brain and DNN. Comparison of the DNN with fMRI RDMs yielded spatial maps of visual representations common to the human brain and the DNN.}
\label{fig:fig2}
\end{figure*}

\begin{figure*}[t]
\begin{center}
   \includegraphics[width=0.80\linewidth]{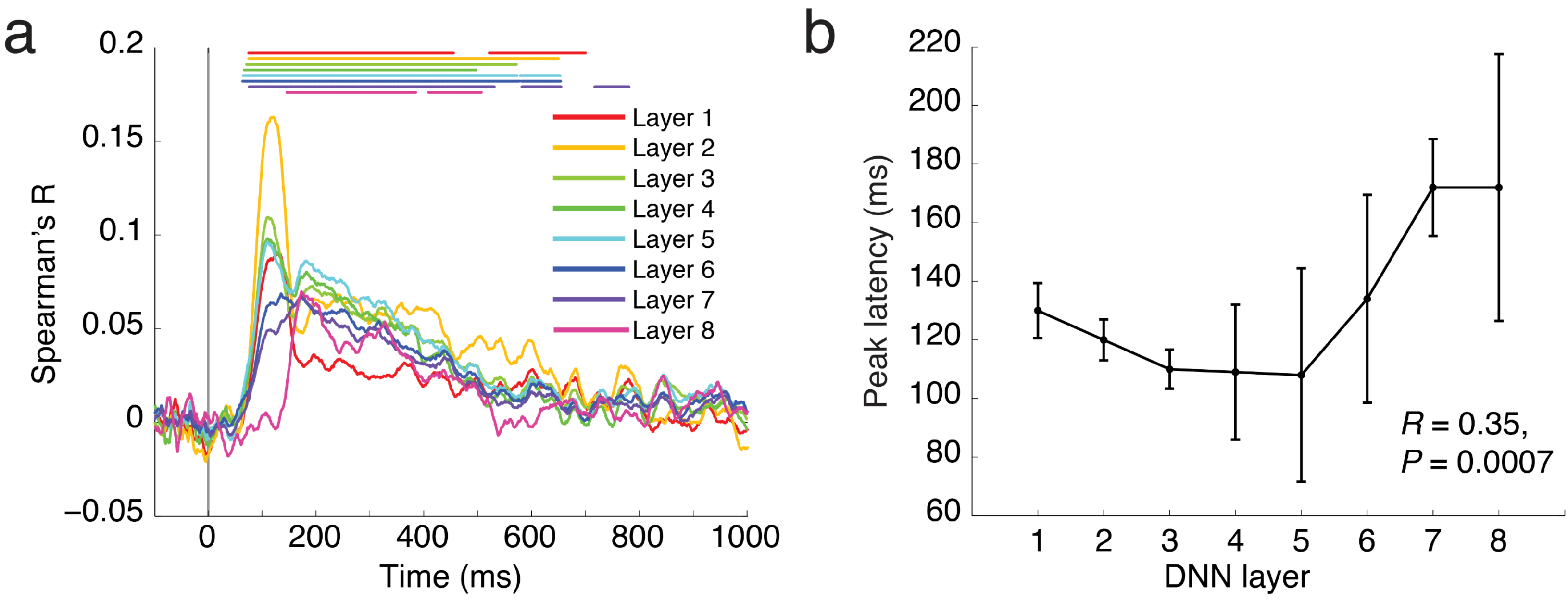}
\end{center}
\vspace*{-6mm}
\caption{\textbf{The object DNN predicted the order of temporally emerging visual representations in the human brain. (a)} Time courses with which representational similarity in the brain and layers of the deep object network emerged. Color-coded lines above data curves indicate significant time points ($n = 15$, cluster definition threshold $P = 0.05$, cluster threshold $P = 0.05$; for onset and peak latencies see Suppl.~Table~2). Gray vertical line indicates image onset. \textbf{(b)} Peak latency of time courses increased with layer number ($n = 15$, $\rho = 0.35$, $P = 0.0007$, sign permutation test), indicating that deeper layers predicted later brain signals. Error bars indicate standard error of the mean determined by 10,000 bootstrap samples of the participant pool.}
\label{fig:fig3}
\vspace*{-2mm}
\end{figure*}

\subsection{Factors determining DNN's predictability of visual representations emerging in time}

The observation of a positive and hierarchical relationship between the object DNN and brain temporal dynamics poses the fundamental question of the origin of this relationship. Three fundamental factors shape DNNs: architecture, task, and training procedure. Determining the effect of each is crucial to understanding the emergence of the brain-DNN relationships on the real-world object categorization task. To this goal, we created several different DNN models (Fig.~\ref{fig:fig5}(a)). We reasoned that a comparison of brain with 1) an untrained DNN would reveal the effect of DNN architecture alone, 2) a DNN trained on an alternate categorization task, scene categorization, would reveal the effect of specific task, and 3) a DNN trained on an image set with random unecological assignment of images to category labels, or a DNN trained on noise images, would reveal the effect of the training procedure per se.

To evaluate the hierarchy of temporal and spatial relationships between the human brain and DNNs, we computed layer-specific RDMs for each DNN. To allow direct comparisons across models, we also computed a single summary RDM for each DNN model based on concatenated layer-specific activation vectors.

Concerning the role of architecture, we found the untrained DNN significantly predicted emerging brain representations (Fig.~\ref{fig:fig5}(b)), but worse than the object DNN (Fig.~\ref{fig:fig5}(c)). A supplementary layer-specific analysis identified every layer as a significant contributor to to this prediction (Suppl.~Fig.~3a). Even though the relationship between layer number and the peak latency of brain-DNN similarity time series was hierarchical, it was negative ($\rho = –0.6, P = 0.0003$, Suppl.~Fig.~3b) and thus reversed and statistically different from the object DNN ($\Delta\rho = 0.96, P = 0.0003$). This shows that DNN architecture alone, independent of task constraints or training procedures, induces representational similarity to emerging visual representations in the brain, but that constraints imposed by training on a real-world categorization task significantly increases this effect and reverses the direction of the hierarchical relationship.

Concerning the role of task, we found the scene DNN also predicted emerging brain representations, but worse than the object DNN (Fig.~\ref{fig:fig5}(b,c); Suppl.~Fig.~3c). This suggests that task constraints influence the model and possibly also brain in a partly overlapping, and partly dissociable manner. Further, the relationship between layer number and brain-DNN similarity time series was positively hierarchical for the scene DNN ($\rho = 0.44, P = 0.001$, Suppl.~Fig.~3(d)), and not different from the object DNN ($\Delta\rho = –0.09, P = 0.41$), further suggesting overlapping neural mechanisms for object and scene perception.

Concerning the role of the training operation, we found both the unecological and noise DNNs predicted brain representations (Fig.~\ref{fig:fig5}(b), Suppl.~Fig~3(e,g)), but worse than the object DNN (Fig.~\ref{fig:fig5}(c)). Further, there was no evidence for a hierarchical relationship between layer number and brain-DNN similarity time series for either DNN (unecological DNN: $\rho = –0.01, P = 0.94$; noise DNN: $\rho = –0.04, P = 0.68$; Suppl.~Fig.~3(f,h)), and both had a weaker hierarchical relationship than the object DNN (unecological DNN: $\Delta\rho = 0.39, P = 0.0107$; noise DNN: $\Delta\rho = 0.36, P = 0.0052$). Thus the training operation per se has an effect on the relationship to the brain, but only training on real-world categorization increases brain-DNN similarity and hierarchy.

In summary, we found that although architecture alone predicted the temporal emergence of visual representations, training on real-world categorization was necessary for a hierarchical relationship to emerge. Thus, both architecture and training crucially influence the prediction power of DNNs over the first few hundred milliseconds of vision.

\subsection{Factors determining DNN's predictability of the topography of visual representations in cortex}
The observation of a positive and hierarchical relationship between the object DNN structure and the brain visual pathways motivates an inquiry, akin to the temporal dynamics analysis in the previous section, regarding the role of architecture, task demands and training operation. For this we systematically investigated three regions-of-interest (ROIs): the early visual area V1, and two regions up-stream in the ventral and dorsal stream, the inferior temporal cortex IT and a region encompassing intraparietal sulcus 1 and 2 (IPS1\&2), respectively. We examined whether DNNs predicted brain activity in these ROIs (Fig.~\ref{fig:fig6}(a)), and also whether this prediction was hierarchical (Fig.~\ref{fig:fig6}, Suppl.~Table~4(a)).

Concerning the role of architecture, we found the untrained DNN predicted brain representations better than the object DNN in V1, but worse in IT and IPS1\&2 (Fig.~\ref{fig:fig6}(a,c)). Further, the relationship was hierarchical (negative) only in IT ($\rho = –0.47, P = 0.002$) (Fig.~\ref{fig:fig6}(b); stars above bars). Thus depending on cortical region the DNN architecture alone is enough to induce similarity between a DNN and the brain, but the hierarchy absent (V1, IPS1\&2) or reversed (IT) without proper DNN training.

Concerning the role of task, we found the scene DNN had largely similar, albeit weaker, similarity to the brain than the object DNN for all ROIs (Fig.~\ref{fig:fig6}(a,c)), with a significant hierarchical relationship in V1 ($\rho = –0.68, P = 0.002$), but not in IT ($\rho = 0.26, P = 0.155$) or IPS1\&2 ($\rho = 0.30, P = 0.08$) (Fig.~\ref{fig:fig6}(b)). In addition, comparing results for the object and scene DNNs directly (Fig.~\ref{fig:fig6}(c)), we found stronger effects for the object DNN in several layers in all ROIs. Together these results corroborate the conclusions of the MEG analysis, showing that task constraints shape brain representations along both ventral visual streams in a partly overlapping, and partly dissociable manner.

Concerning the role of the training operation, we found both the unecological and noise DNNs predicted visual representations in V1 and IT, but not IPS1\&2 (Fig.~\ref{fig:fig6}(a)), and with less predictive power than the object DNN in all regions (Fig.~\ref{fig:fig6}(c)). A hierarchical relationship was present and negative in V1 and IT, but not IPS1\&2 (Fig.~\ref{fig:fig6}(b), unecological DNN: V1 $\rho = –0.40, P = 0.001$, IT $\rho = –0.38, P = 0.001$, IPS1\&2 $\rho = –0.03, P = 0.77$; noise DNN: V1 $\rho= –0.08, P = 0.42$, IT $\rho = –0.29, P =0.012$, IPS1\&2 $\rho=–0.08, P = 0.42$).

Therefore the training on a real-world categorization task, but not the training operation per se, increases the brain-DNN similarity while inducing a hierarchical relationship.

\begin{figure*}[t]
\begin{center}
   \includegraphics[width=0.83\linewidth]{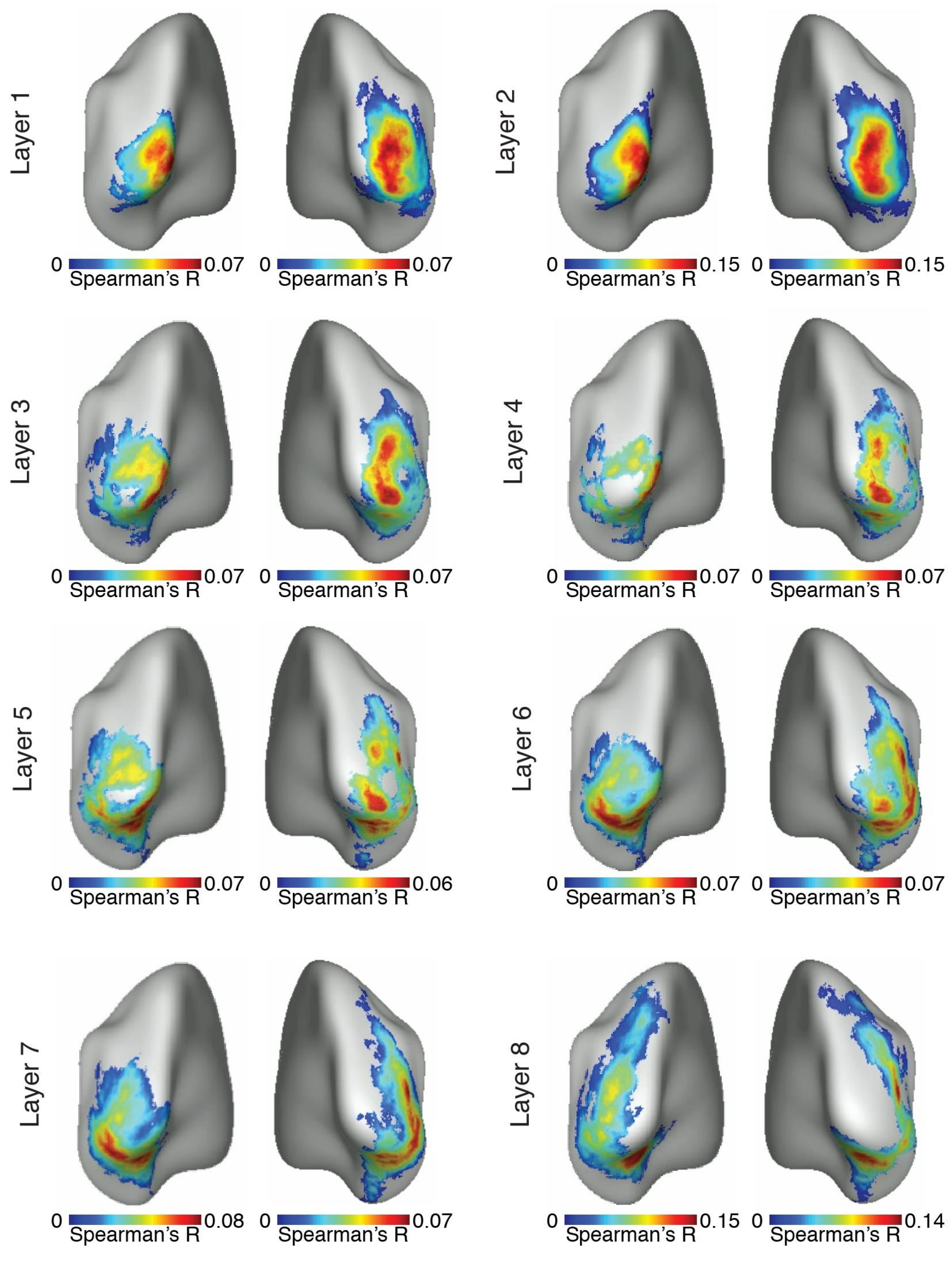}
\end{center}
\vspace*{-2mm}
\caption{\textbf{Spatial maps of visual representations common to brain and object DNN.} The object DNN predicted the hierarchical topography of visual representations in the human brain. Low layers had significant representational similarities confined to the occipital lobe of the brain, i.e. low- and mid-level visual regions. Higher layers had significant representational similarities with more anterior regions in the temporal and parietal lobe, with layers 7 and 8 reaching far into the inferior temporal cortex and inferior parietal cortex ($n = 15$, cluster definition threshold $P < 0.05$, cluster-threshold $P < 0.05$, analysis separate for each hemisphere).}
\label{fig:fig4}
\end{figure*}

\begin{figure*}[t]
\begin{center}
   \includegraphics[width=0.80\linewidth]{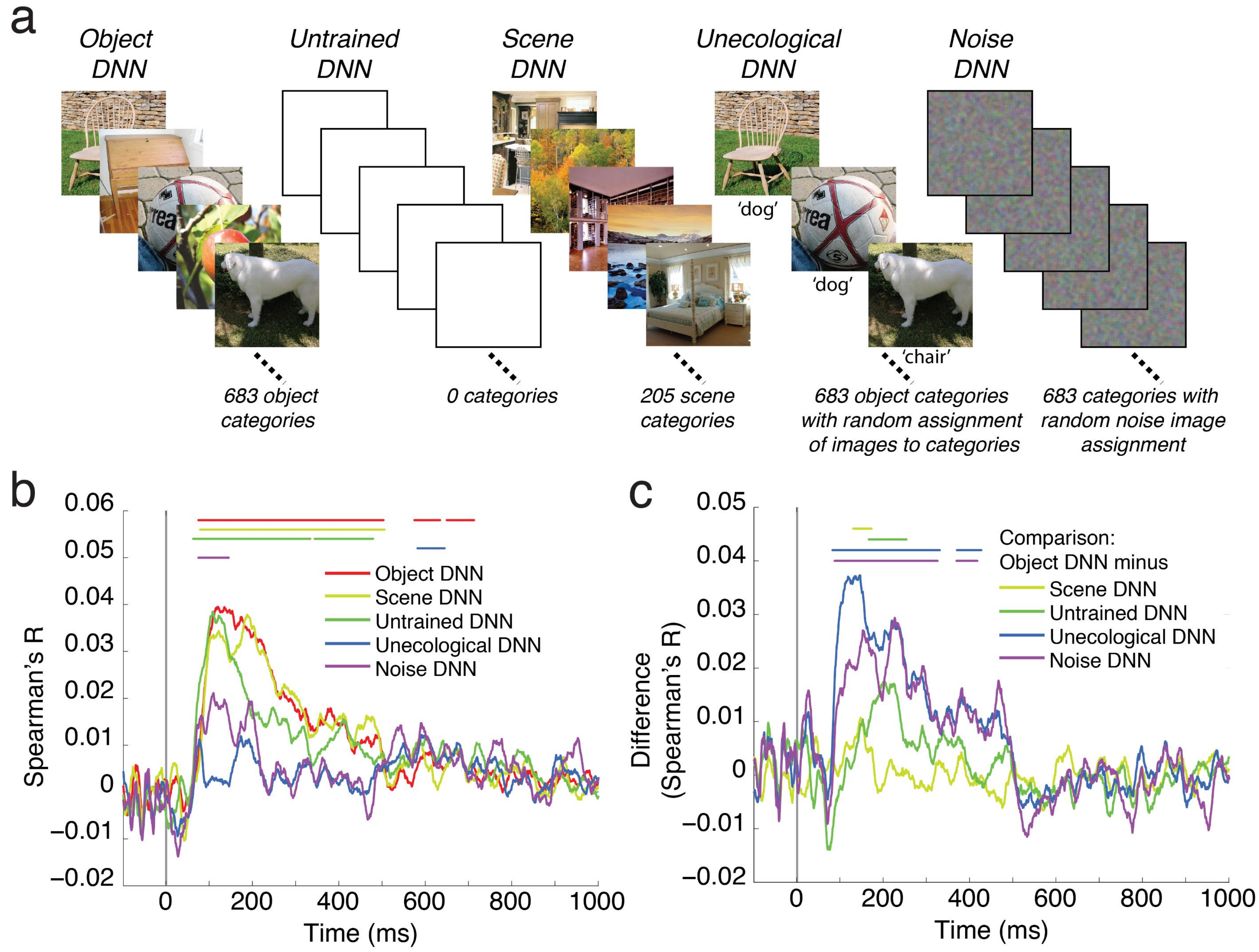}
\end{center}
\vspace*{-6mm}
\caption{\textbf{Architecture, task, and training procedure influence the DNN's predictability of temporally emerging brain representations. (a)} We created 5 different models: (1) a model trained on object categorization (\textit{object DNN}; Fig.~\ref{fig:fig1}); (2) an untrained model initialized with random weights (\textit{untrained DNN}) to determine the effect of architecture alone; (3) a model trained on a different real-world task, scene categorization (\textit{scene DNN}) to investigate the effect of task; and (4,5) a model trained on object categorization with random assignment of image labels (\textit{unecological DNN}), or spatially smoothed noisy images with random assignment of image labels (\textit{noise DNN}), to determine the effect of the training operation independent of task constraints. \textbf{(b)} All DNNs had significant representational similarities to human brains (layer-specific analysis in Suppl.~Fig.~3). \textbf{(c)} We contrasted the object DNN against all other models (subtraction of corresponding time series shown in (b)). Representations in the object DNN were more similar to brain representations than any other model, though the scene DNN was a close second. Lines above data curves significant time points ($n = 15$, cluster definition threshold $P = 0.05$, cluster threshold $P = 0.05$; for onset and peak latencies see Suppl.~Table~3(a,b)). Gray vertical lines indicates image onset.}
\label{fig:fig5}
\end{figure*}

\section{Discussion}
By comparing the spatio-temporal dynamics in the human brain with a deep neural network (DNN) model trained on object categorization, we provided a formal model of object recognition in cortex. We found a correspondence between the object DNN and the brain in both space (fMRI data) and time (MEG data). Both cases demonstrated a hierarchy: in space from low- to high-level visual areas in both ventral and dorsal stream, in time over the visual processing stages in the first few hundred milliseconds of vision. A systematic analysis of the fundamental determinants of this DNN-brain relationship identified that the architecture alone induces similarity, but that training on a real-world categorization task was necessary for a hierarchical relationship to emerge. Our results demonstrate the explanatory and discovery power of the brain-DNN comparison approach to understand the spatio-temporal neural dynamics underlying object recognition. They provide novel evidence for a role of parietal cortex in visual object categorization, and give rise to the idea that the organization of the visual cortex may be influenced by processing constraints imposed by visual categorization the same way that DNN representations were influenced by object categorization tasks.

\subsection{Object DNN predicts a hierarchy of brain representations in space and time}
A major impediment in modeling human object recognition in cortex is the lack of principled understanding of exact neuronal tuning in mid- and high-level visual cortex. Previous approaches thus extrapolated principles observed in low-level visual cortex, with limited success in capturing neuronal variability and a much inferior to human behavioral performance~\cite{riesenhuber1999hierarchical,riesenhuber2002neural}.

Our approach allowed us to obviate this limitation by relying on an object recognition model that learns neuronal tuning. By comparing representations between the DNN and the human brain we found a hierarchical correspondence in both space and time: early layers of the DNN predicted visual representations emerging early after stimulus onset, and in regions low in the cortical processing hierarchy, with progressively higher DNN layers predicting subsequent emerging representations in higher regions of both the dorsal and ventral visual pathway. Our results provide algorithmically informed evidence for the idea of visual processing as a step-wise hierarchical process in time~\cite{bullier2001integrated,cichy2014resolving,mormann2008latency} and along a system of cortical regions~\cite{felleman1991distributed,dicarlo2012does,freiwald2009face}.

In regards to the temporal correspondence in particular, our results provide first evidence for a hierarchical relationship between computer models of vision and the brain. Peak latencies between layers of the object DNN and emerging brain activations ranged between approximately $100$ and $160ms$. While in agreement with prior findings about the time necessary for complex object processing~\cite{thorpe1996speed}, our results go further by making explicit the step-wise transformations of representational format that may underlie rapid complex object categorization behavior.

In regards to the spatial correspondence, previous studies compared DNNs to the ventral visual stream only, mostly using a spatially limited region-of-interest approach ~\cite{gucclu2014deep,khaligh2014deep,yamins2014performance}. Here, using a spatially unbiased whole-brain approach~\cite{kriegeskorte2006information}, we discovered a hierarchical correspondence in the dorsal visual pathway. While previous studies have documented object selective responses in dorsal stream in monkeys~\cite{janssen2008coding,sawamura2005using} and humans~\cite{chao2000representation,konen2008two}, it is still debated whether dorsal visual representations are better explained by differential motor action associations or ability to engage attention, rather than category membership or shape representation~\cite{grill1999differential,kourtzi2000cortical}. Crucially, our results defy explanation by attention or motor-related concepts, as neither played any role in the DNN and thus brain-DNN correspondence. Concurrent with the observation that temporal lobe resection shows limited behavioral effect in object recognition~\cite{buckley1997functional,weiskrantz1984impairments}, our results argue that parietal cortex might play a stronger role in object recognition than previously appreciated.

Our results thus challenge the classic descriptions of the dorsal pathway as a spatially- or action oriented `where' or `how' pathway~\cite{goodale1982analysis,milner1995visual}, and suggest that current theories describing parietal cortex as related to spatial working memory, visually guided actions and spatial navigation~\cite{kravitz2011new} should be complemented with a role for the dorsal visual stream in object categorization~\cite{konen2008two}.

\begin{figure*}[t]
\begin{center}
   \includegraphics[width=\linewidth]{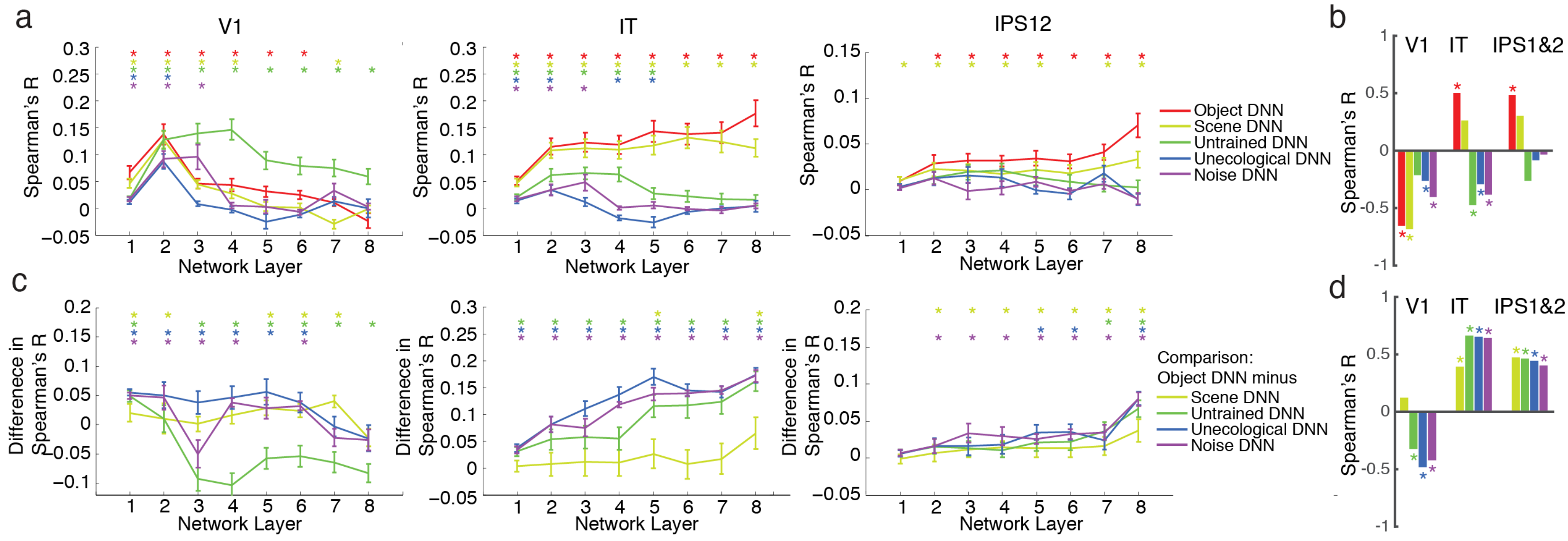}
\end{center}
\vspace*{-4mm}
\caption{\textbf{Architecture, task constraints, and training procedure influence the DNN's predictability of the topography of brain representations. (a)} Comparison of fMRI representations in V1, IT and IPS1\&2 with the layer-specific DNN representations of each model. Error bars indicate standard error of the mean as determined by bootstrapping ($n = 15$). \textbf{(b)} Correlations between layer number and brain-DNN representational similarities for the different models shown in (a). Non-zero correlations indicate hierarchical relationships; positive correlations indicate an increase in brain-DNN similarities towards higher layers, and vice versa for negative correlations. Bars color-coded as DNNs, stars above bars indicate significance (sign-permutation tests, $P < 0.05$, FDR-corrected, for details see Suppl.~Table~4(a)). \textbf{(c)} Comparison of object DNN against all other models (subtraction of corresponding points shown in (a)). \textbf{(d)} Same as (b), but for the curves shown in (c) (for details see Suppl.~Table~4b).}
\label{fig:fig6}
\end{figure*}

\subsection{Origin and implications of brain-DNN representation similarities}
Investigating the influence of crucial parameters determining DNNs, we found an influence of both architecture and task constraints induced by training the DNN on a real-world categorization task. This suggests that that similar architectural principles, i.e., convolution, max pooling and normalization govern both model and brains, concurrent with the origin of those principle by observation in the brain~\cite{riesenhuber1999hierarchical}. The stronger similarity with early rather than late brain regions might be explained by the fact that neural networks initialized with random weights that involve a convolution, nonlinearity and normalization stage exhibit Gabor-like filters sensitive to oriented edges, and thus similar properties an neurons in early visual areas~\cite{saxe2011random}.

Although architecture alone induced similarity, training on a real-world categorization tasks increased similarity and was necessary for a hierarchical relationship in processing stages between the brain and the DNN to emerge in space and time. This demonstrates that learning constraints imposed by a real-world categorization task crucially shape the representational space of a DNN~\cite{yamins2014performance}, and suggests that the processing hierarchy in the human brain is a the result of computational constraints imposed by visual object categorization. Such constraints may originate in high-level visual regions such as IT and IPS, be propagated backwards from high-level visual regions through the visual hierarchies through abundantly present feedback connections in the visual stream at all levels~\cite{van1994multiple} during visual learning~\cite{ahissar2004reverse}, and provide the basis of learning at all stages of the processing in visual brain~\cite{kourtzi2006learning}.

\subsection{Summary statement}
In sum, by comparing deep neural networks to human brains in space and time, we provide a spatio-temporally unbiased algorithmic account of visual object recognition in human cortex.

\section{Method}
\textbf{Participants:} 
15 healthy human volunteers (5 female, age: mean $\pm$ s.d. $= 26.6 \pm 5.18$ years, recruited from a subject pool at Massachusetts Institute of Technology) participated in the experiment. The sample size was based on methodological recommendations in literature for random-effects fMRI and MEG analyses. Written informed consent was obtained from all subjects. The study was approved by the local ethics committee (Institutional Review Board of the Massachusetts Institute of Technology) and conducted according to the principles of the declaration of Helsinki. All methods were carried out in accordance with the approved guidelines.

\textbf{Visual stimuli:}
The stimuli presented to humans and computer vision models were 118 color photographs of everyday objects, each from a different category, on natural backgrounds (Fig.~\ref{fig:fig2}(b)) from the ImageNet database~\cite{deng2009imagenet}.

\subsection{Experimental design and task}
Participants viewed images presented at the center of the screen ($4\degree$ visual angle) for $0.5s$ and overlaid with a light gray fixation cross. The presentation parameters were adapted to the specific requirements of each acquisition technique (Suppl.~Fig.~1).

For MEG, participants completed $15$ runs of $314s$ duration. Each image was presented twice in each MEG run in random order with an inter-trial interval (ITI) of $0.9-1s$. Participants were asked to press a button and blink their eyes in response to a paper clip image shown randomly every 3 to 5 trials (average 4). The paper clip image was not part of the image set, and paper clip trials were excluded from further analysis.

For fMRI, each participant completed two independent sessions of $9-11$ runs ($486 s$ duration each) on two separate days. Each run consisted of one presentation of each image in random order, interspersed randomly with 39 null trials (i.e., $25\%$ of all trials) with no stimulus presentation. During the null trials the fixation cross turned darker for $500ms$. Participants reported changes in fixation cross hue with a button press.

\newpage
\textbf{MEG acquisition:}
MEG signals were acquired continuously from 306 channels (204 planar gradiometers, 102 magnetometers, Elektra Neuromag TRIUX, Elekta, Stockholm) at a sampling rate of $1000$Hz, and filtered online between $0.03$ and 330Hz. We preprocessed data with temporal source space separation (maxfilter software, Elekta, Stockholm) before further analysis with Brainstorm\footnote{\href{http://neuroimage.usc.edu/brainstorm/}{http://neuroimage.usc.edu/brainstorm/}}. We extracted each trial with a $100ms$ baseline and $1000ms$ post-stimulus recordings, removed baseline mean, smoothed data with a 30Hz low-pass filter, and normalized each channel with its baseline standard deviation. This yielded 30 preprocessed trials per condition and participant.

\textbf{fMRI acquisition:}
Magnetic resonance imaging (MRI) was conducted on a 3T Trio scanner (Siemens, Erlangen, Germany) with a 32-channel head coil. We acquired structural images using a standard T1-weighted sequence (192 sagittal slices, FOV = $256 mm^2$, TR $= 1900ms$, TE = $2.52 ms$, flip angle $= 9\degree$).

For fMRI, we conducted $9–11$ runs in which 648 volumes were acquired for each participant (gradient-echo EPI sequence: TR $= 750 ms$, TE $= 30 ms$, flip angle $= 61\degree$, FOV read $= 192 mm$, FOV phase $= 100\%$ with a partial fraction of $\frac{6}{8}$, through-plane acceleration factor 3, bandwidth 1816Hz/Px, resolution = $3mm^3$, slice gap $20\%$, slices $= 33$, ascending acquisition). The acquisition volume covered the whole cortex.

\subsection{Anatomical MRI analysis}
We reconstructed the cortical surface of each participant using Freesurfer on the basis of the T1 structural scan~\cite{dale1999cortical}. This yielded a discrete triangular mesh representing the cortical surface used for the surface-based two-dimensional (2D) searchlight procedure outlined below.

\textbf{fMRI analysis:}
We preprocessed fMRI data using SPM8\footnote{\href{http://www.fil.ion.ucl.ac.uk/spm/}{http://www.fil.ion.ucl.ac.uk/spm/}}. For each participant and session separately, fMRI data were realigned and co-registered to the T1 structural scan acquired in the first MRI session. Data was neither normalized nor smoothed. We estimated the fMRI response to the 118 image conditions with a general linear model. Image onsets and duration were entered into the GLM as regressors and convolved with a hemodynamic response function. Movement parameters entered the GLM as nuisance regressors. We then converted each of the 118 estimated GLM parameters into $t$-values by contrasting each condition estimate against the implicitly modeled baseline. Additionally, we determined the grand-average effect of visual stimulation independent of condition in a separate t-contrast of parameter estimates for all 118 image conditions versus the implicit baseline.

\textbf{Definition of fMRI regions of interest:}
We defined three regions-of-interest for each participant: V1 corresponding to the central 4° of the visual field, inferior temporal cortex (IT), and intraparietal sulcus regions 1 and 2 combined (IPS1\&2). We defined the V1 ROI based on an anatomical eccentricity template~\cite{benson2012retinotopic}. For this, we registered a generic V1 eccentricity template to reconstructed participant-specific cortical surfaces and restricted the template to the central 4° of visual angle. The surface-based ROIs for the left and right hemisphere were resampled to the space of EPI volumes and combined.

To define inferior temporal cortex (IT), we used an anatomical mask of bilateral fusiform and inferior temporal cortex (WFU Pickatlas, IBASPM116 Atlas). To define IPS1\&2, we used a combined probabilistic mask of IPS1 and IPS2~\cite{wang2014probabilistic}. Masks in MNI space were reverse-normalized to single-subject functional space. We then restricted the anatomical definition of each ROI for each participant by functional criteria to the 100 most strongly activated voxels in the grand-average contrast of visual stimulation vs. baseline.

\textbf{fMRI surface-based searchlight construction and analysis:}
To analyze fMRI data in a spatially unbiased (unrestricted from ROIs) approach, we performed a 2D surface-based searchlight analysis following the approach of Chen et al~\cite{chen2011cortical}. We used a cortical surface-based instead of a volumetric searchlight procedure as the former promises higher spatial specificity. The construction of 2D surface-based searchlights was a two-point procedure. First, we defined 2D searchlight disks on subject-specific reconstructed cortical surfaces by identifying all vertices less than $9mm$ away in geodesic space for each vertex v. Geodesic distances between vertices were approximated by the length of the shortest path on the surface between two vertices by Dijkstra's algorithm~\cite{dale1999cortical}. Second, we extracted fMRI activity patterns in functional space corresponding to the vertices comprising the searchlight disks. Voxels belonging to a searchlight were constrained to appear only once in a searchlight, even if they were nearest neighbor to several vertices. For random effects analysis, i.e., to summarize results across subjects, we estimated a mapping between subject-specific surfaces and an average surface using freesurfer~\cite{dale1999cortical} (fsaverage).

\subsection{Convolutional neural network architecture and training}

We used a deep neural network (DNN) architecture as described by Krizhevsky et al~\cite{krizhevsky2012imagenet} (Fig.~\ref{fig:fig1}(a)). We chose this architecture because it was the best-performing neural network in the ImageNet Large Scale Visual Recognition Challenge 2012, it is inspired by biological principles. The network architecture consisted of 8 layers; the first five layers were convolutional; the last three were fully connected. Layers 1 and 2 consisted of three stages: convolution, max pooling and normalization; layers $3-5$ consisted of a convolution stage only (enumeration of units and features for each layer in Suppl.~Table~5). We used the last processing stage of each layer as model output of each layer for comparison with fMRI and MEG data.

We constructed 5 different DNN models that differed in the categorization task they were trained on (Fig.~\ref{fig:fig5}(a)): (1) object DNN, i.e., a model trained on object categorization; (2) untrained DNN, i.e., an untrained model initialized with random weights; (3) scene DNN, i.e., a model trained on scene categorization; (4) unecological DNN, i.e., a model trained on object categorization but with random assignment of label to the training image set; and (5) noise DNN, i.e., a model trained to categorize structured noise images. In detail, the object DNN was trained with 900k images of 683 different objects from ImageNet~\cite{deng2009imagenet} with roughly equal number of images per object ($\sim1300$). The scene DNN, was trained with the recently released Places dataset that contains images from different scene categories~\cite{zhou2014learning}. We used 216 scene categories and normalized the total number of images to be equivalent to the number of images used to train the object DNN. For the noise DNN we created an image set consisting of 1000 random categories of 1300 images each. All noise images were sampled independently of each other and had size $256\times256$ with 3 color channels. To generate, each color channel and pixel was sampled independently from a uniform $[0,1]$ distribution, followed by convolution with a 2D Gaussian filter of size $10\times10$ with standard deviation of 80 pixels. The resulting noise images had small but perceptible spatial gradients.

All DNNs except the untrained DNN were trained on GPUs using the Caffe toolbox\footnote{\href{http://caffe.berkeleyvision.org/}{http://caffe.berkeleyvision.org/}} with the learning parameters set as follows: the networks were trained for 450k iterations, with the initial learning rate set to 0.01 and a step multiple of 0.1 every 100k iterations. The momentum and weight decay were fixed at 0.9 and 0.0005 respectively.

To ascertain that we successfully trained the networks, we determined their performance in predicting the category of images in object and scene databases based on the output of layer 7. As expected, the deep object- and scene networks performed comparably to previous DNNs trained on object and scene categorization, whereas the unecological and noise networks performed at chance level (Suppl.~Table~1).

To determine classification accuracy of the object DNN on the 118-image set used to probe the brain here, we determined the 5 most confident classification labels for each image. We then manually verified whether the predicted labels matched the expected object category. Manual verification was required to correctly identify categories that were visually very similar but had different labels e.g., backpack and book bag, or airplane and airliner. Images belonging to categories for which the network was not trained (i.e., person, apple, cattle, sheep) were marked as incorrect. Overall, the network classified $111/118$ images correctly, resulting in a $94\%$ success rate, comparable to humans~\cite{russakovsky2014imagenet} (image-specific voting results available online at \href{http://brainmodels.csail.mit.edu}{http://brainmodels.csail.mit.edu}).

\subsection{Visualization of model neuron receptive field properties and DNN connectivity}
We used a neuroscience-inspired reduction method to determine the receptive field (RF) properties size and selectivity of model neurons~\cite{scenecnn_iclr15}. In short, for any neuron we determined the $K = 25$ most-strongly activating images. To determine the empirical size of the RF, we replicated the $K$ images many times with small random occluders at different positions in the image. We then passed the occluded images into the DNN and compared the output to the original image, thus constructing a discrepancy map that indicates which portion of the image drives the neuron. Re-centering and averaging discrepancy maps generated the final RF.

To illustrate the selectivity of neuron RFs, we use shaded regions to highlight the image area primarily driving the neuron response (Fig.~\ref{fig:fig1}(b)). This was obtained by first producing the neuron feature map (the output of a neuron to a given image as it convolves the output of the previous layer), then multiplying the neuron RF with the value of the feature map in each location, summing the contribution across all pixels, and finally thresholding this map at $50\%$ of its maximum value.

To illustrate the parameters of the object deep network, we developed a tool (DrawNet; \href{http://brainmodels.csail.mit.edu}{http://brainmodels.csail.mit.edu}) that plots for any chosen neuron in the model 1) the selectivity of the neuron for a particular image, and the strongest connections (weights) between the neurons in the previous and next layer. Only connections with weights that exceed a threshold of 0.75 times the maximum weight for a particular neuron are displayed. DrawNet plots properties for the pooling stage of layers 1, 2 and 5 and for the convolutional stage of layers 3 and 4.

\subsection{Analysis of fMRI, MEG and computer model data in a common framework}
To compare brain imaging data (fMRI, MEG) with the DNN in a common framework we used representational similarity analysis~\cite{khaligh2014deep,kriegeskorte2008representational}. The basic idea is that if two images are similarly represented in the brain, they should be similarly represented in the computer model, too. Pair-wise similarities, or equivalently dissimilarities, between the 118 condition-specific representations can be summarized in a representational dissimilarity matrix (RDM) of size $118\times118$, indexed in rows and columns by the compared conditions. Thus representational dissimilarity matrices can be calculated for fMRI (one fMRI RDM for each ROI or searchlight), for MEG (one MEG RDM for each millisecond), and for DNNs (one DNN RDM for each layer). In turn, layer-specific DNN RDMs can be compared to fMRI or MEG RDMs yielding a measure of brain-DNN representational similarity. The specifics of RDM construction for MEG, fMRI and DNNs are given below.

\subsection{Multivariate analysis of fMRI data yields space-resolved fMRI representational dissimilarity matrices}
To compute fMRI RDMs we used a correlation-based approach. The analysis was conducted independently for each subject. First, for each ROI (V1, IT, or IPS1\&2) and each of the 118 conditions we extracted condition-specific $t$-value activation patterns and concatenated them into vectors, forming 118 voxel pattern vectors of length V=100. We then calculated the dissimilarity ($1 -$ Spearman's $\rho$) between $t$-value patterns for every pair of conditions. This yielded a $118\times118$ fMRI representational dissimilarity matrix (RDM) indexed in rows and columns by the compared conditions for each ROI. Each fMRI RDM was symmetric across the diagonal, with entries bounded between 0 (no dissimilarity) and 2 (complete dissimilarity).

To analyze fMRI data in a spatially unbiased fashion we used a surface-based searchlight method. Construction of fMRI RDMs was similar to the ROI case above, with the only difference that activation pattern vectors were formed separately for each voxel by using $t$-values within each corresponding searchlight, thus resulting in voxel-resolved fMRI RDMs.

\subsection{Construction of DNN layer-resolved and summary DNN representational dissimilarity matrices}
To compute DNN RDMs we again used a correlation-based approach. For each layer of the DNN, we extracted condition-specific model neuron activation values and concatenated them into a vector. Then, for each condition pair we computed the dissimilarity ($1-$ Spearman's $\rho$) between the model activation pattern vectors. This yielded a $118\times118$ DNN representational dissimilarity matrix (DNN RDM) summarizing the representational dissimilarities for each layer of a network. The DNN RDM is symmetric across the diagonal and bounded between 0 (no dissimilarity) and 2 (complete dissimilarity).

For an analysis of representational dissimilarity at the level of whole DNNs rather than individual layers we modified the aforementioned procedure (Fig.~\ref{fig:fig5}(b)). Layer-specific model neuron activation values were concatenated before entering similarity analysis, yielding a single DNN RDM per model. To balance the contribution of each layer irrespective of the highly different number of neurons per layer, we applied a principal component analysis (PCA) on the condition- and layer-specific activation patterns before concatenation, yielding 117-dimensional summary vectors for each layer and condition. Concatenating the 117-dimensional vector across 8 layers yielded a $117\times8=936$ dimensional vector per condition that entered similarity analysis.

\subsection{Multivariate analysis of MEG data yields time-resolved MEG representational dissimilarity matrices}
To compute MEG RDMs we used a decoding approach with a linear support vector machine (SVM). The idea is that if a classifier performs well in predicting condition labels based on MEG data, then the MEG visual representations must be sufficiently dissimilar. Thus, decoding accuracy of a classifier can be interpreted as a dissimilarity measure. The motivation for a classifier-based dissimilarity measure rather than $1-$ Spearman's $\rho$ (as above) is that a SVM selects MEG sensors that contain discriminative information in noisy data without human intervention. A dissimilarity measure over all sensors might be strongly influences by noisy channels, and an a-priori sensor selection might introduce a bias, and neglect the fact that different channels contain discriminate information over time.

We extracted MEG sensor level patterns for each millisecond time point ($100ms$ before to $1000ms$ after image onset) and for each trial. For each time point, MEG sensor level activations were arranged in 306 dimensional vectors (corresponding to the 306 MEG sensors), yielding $M=30$ pattern vectors per time point and condition). To reduce computational load and improve signal-to-noise ratio, we sub-averaged the $M$ vectors in groups of $k = 5$ with random assignment, thus obtaining $L=M/k$ averaged pattern vectors. For each pair of conditions, we assigned $L-1$ averaged pattern vectors to a training data set used to train a linear support vector machine in the LibSVM implementation\footnote{\href{http://www.csie.ntu.edu.tw/~cjlin/libsvm}{http://www.csie.ntu.edu.tw/$\sim$cjlin/libsvm}}. The trained SVM was then used to predict the condition labels of the left-out testing data set consisting of the $L^{\text{th}}$ averaged pattern vector. We repeated this process 100 times with random assignment of the $M$ raw pattern vectors to $L$ averaged pattern vectors. We assigned the average decoding accuracy to a decoding accuracy matrix of size $118 \times 118$, with rows and columns indexed by the classified conditions. The matrix was symmetric across the diagonal, with the diagonal undefined. This procedure yielded one $118 \times 118$ matrix of decoding accuracies and thus one MEG representational dissimilarity matrix (MEG RDM) for every time point.

\subsection{Representational similarity analysis compares brain data to DNNs}
We used representational similarity analysis to compare layer-specific DNN RDMs to space-resolved fMRI RDMs or time-resolved MEG RDMs (Fig.~\ref{fig:fig2}(b)). In particular, fMRI or MEG RDMs were compared to layer-specific DNN RDMs by calculating Spearman's correlation between the lower half of the RDMs excluding the diagonal. All analyses were conducted on single-subject basis. 

A comparison of time-resolved MEG RDMs and DNN RDMs (Fig.~\ref{fig:fig2}(b)) yielded the time course with which visual representations common to brains and DNNs emerged. For the comparison of fMRI and DNNs RDMs, fMRI searchlight (Fig.~\ref{fig:fig2}(b)) and ROI RDMs were compared with DNN RDMs, yielding single ROI values and 2-dimensional brain maps of similarity between human brains and DNNs respectively. 

For the searchlight-based fMRI-DNN comparison procedure in detail, we computed the Spearman's $\rho$ between the DNN RDM of a given layer and the fMRI RDM of a particular voxel in the searchlight approach. The resulting similarity value was assigned to a 2D map at the location of the voxel. Repeating this procedure for each voxel yielded a spatially resolved similarity map indicating common brain-DNN representations. The entire analysis yielded 8 maps, i.e., one for each DNN layer. Subject-specific similarity maps were transformed into a common average cortical surface space before entering random-effects analysis.

\subsection{Statistical testing}
For random-effects inference we used sign permutation tests. In short, we randomly changed the sign of the data points (10,000 permutation samples) for each subject to determine significant effects at a threshold of $P < 0.05$. To correct for multiple comparisons in cases where neighboring tests had a meaningful structure, i.e., neighboring voxels in the searchlight analysis and neighboring time points in the MEG analysis, we used cluster-size inference with a cluster-size threshold of $P < 0.05$. In other cases, we used FDR correction.

To provide estimates of the accuracy of a statistic we bootstrapped the pool of subjects (1000 bootstraps) and calculated the standard deviation of the sampled bootstrap distribution. This provided the standard error of the statistic.
 

\section*{Acknowledgements}
We thank Chen Yi for assisting in surface-based searchlight analysis. This work was funded by National Eye Institute grant EY020484 (to A.O.), a Google Research Faculty Award (to A.O.), a Feodor Lynen Scholarship of the Humboldt Foundation (to R.M.C), the McGovern Institute Neurotechnology Program (to A.O. and D.P.), and was conducted at the Athinoula A. Martinos Imaging Center at the McGovern Institute for Brain Research, Massachusetts Institute of Technology.


\section*{Author Contributions}
All authors conceived the experiments. R.M.C. and D.P. acquired and analyzed brain data, A.K. trained and analyzed computer models. R.M.C. provided model-brain comparison. R.M.C., A.K., D.P. and A.O. wrote the paper, A.T. provided expertise and feedback. A.O., D.P. and R.M.C. provided funding.

\section*{Competing Interests Statement}
The authors have no competing financial interests.

{\small
\bibliographystyle{ieee}
\bibliography{main}
}

\end{document}